\documentclass[10pt,twocolumn,letterpaper]{article}

\usepackage[preprint]{cvpr}      
\usepackage{dirtytalk}
\usepackage{makecell}
\usepackage{multirow}
\usepackage{xcolor}
\definecolor{cvprblue}{rgb}{0.21,0.49,0.74}
\usepackage[pagebackref,breaklinks,colorlinks,allcolors=cvprblue]{hyperref}

\def\confName{CVPR}
\def\confYear{2027}

\title{Lossless Latent Diffusion Models}

\author{Tongda Xu, Yu Zhang, Jun Hu, Huijuan Huang\\
Apple\\
{\tt\small x.tongda@nyu.edu}
}

\begin{document}
\twocolumn[{
\maketitle
\begin{center}
    \captionsetup{type=figure}
    \centering    \includegraphics[width=\textwidth]{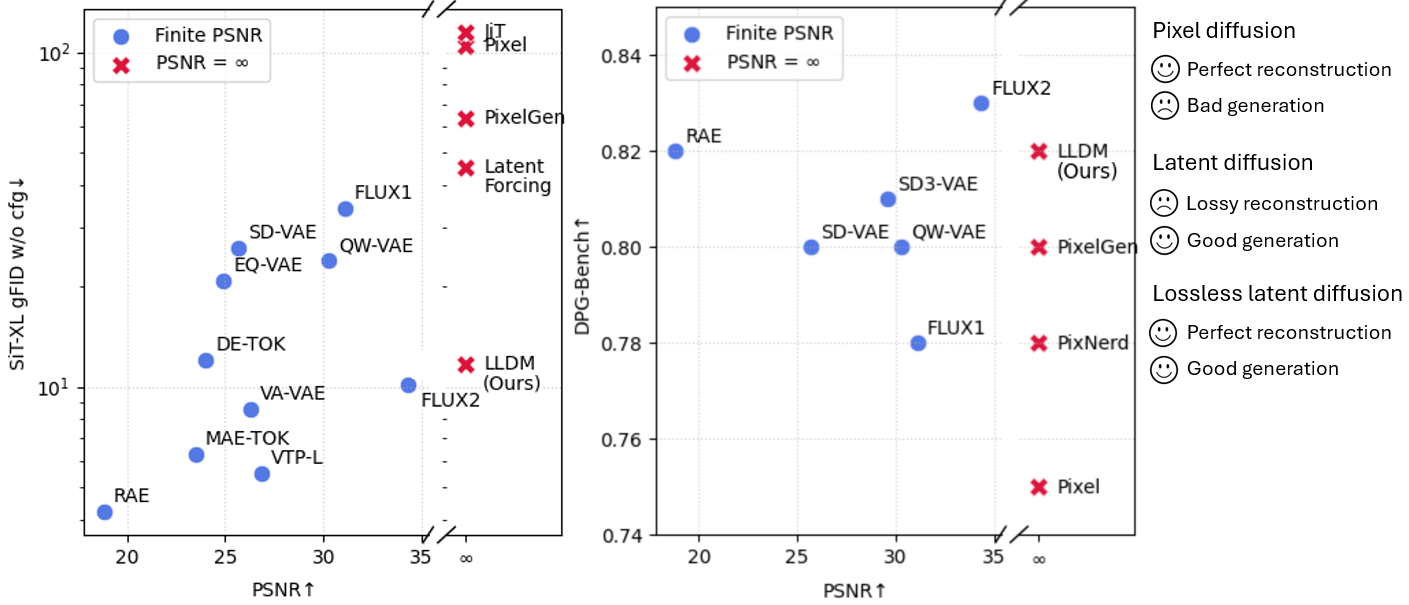}
    \captionof{figure}{Lossless LDM achieves pixel-exact reconstruction (PSNR$=\infty$) while outperforming many lossy LDMs in generation quality. \textit{Left:} ImageNet class-to-image generation. \textit{Right:} Diffusion-Bench text-to-image generation.}
        \label{fig:cov}
\end{center}
}]

\begin{abstract}

We study the problem of training image diffusion models while preserving pixel-exact reconstruction. Prior works typically address this problem by directly modeling images in pixel space, requiring modifications to the diffusion training objective or model architecture. In this work, we introduce a different paradigm: rather than modifying diffusion models, \textbf{we construct a lossless transformation that maps pixels into a representation more amenable to diffusion modeling}. We term our approach the Lossless Latent Diffusion Model (LLDM). We construct LLDM using residual latent transformation and uneven latent scaling, and explain its effectiveness through a proposed metric, semantic rank priority (SRP). Empirically, LLDM substantially outperforms prior pixel-space diffusion models in generation quality and even surpasses several widely used lossy VAEs. Furthermore, we show that SRP is strongly correlated with downstream generative performance.
\end{abstract}

\section{Introduction}

Training diffusion models directly in pixel space is appealing because it avoids the reconstruction errors introduced by variational autoencoders (VAEs). However, pixel-space diffusion is shown to be challenging for high-quality image generation, partly due to the high intrinsic dimensionality of the pixel-space data manifold~\citep{li2026back,Jiang2026AnES}. Recent advances in pixel-space diffusion have primarily addressed this challenge by modifying either the training procedure or the diffusion model itself. The former introduces auxiliary objectives, such as finetuning from latent diffusion \citep{Jiang2026AnES}, $x$-prediction~\citep{li2026back}, perceptual losses~\citep{Ma2026PixelGenIP}, and DINO feature losses~\citep{Baade2026LatentFR}. The latter develops architectures tailored to pixel-space modelling, including pixel-field parameterization~\citep{Wang2025PixNerdPN} and hierarchical architectures~\citep{Yu2025PixelDiTPD,Teng2023RelayDU,Ma2025DeCoFP}.

In contrast, latent diffusion models (LDMs)~\citep{Rombach2021HighResolutionIS} simplify diffusion modeling by mapping images into lower-dimensional representations using lossy VAEs. This improves modeling efficiency at the cost of reconstruction fidelity. Early LDM research therefore focused on improving VAE reconstruction quality~\citep{Esser2024ScalingRF}. However, recent studies have identified a \say{reconstruction--generation trade-off}: better reconstruction fidelity does not necessarily translate into better downstream generation quality~\citep{Yao2025ReconstructionVG}. This has motivated efforts to learn more diffusion-friendly representations through foundation-model alignment or signal-processing-based transformations~\citep{Yu2024RepresentationAF,Zheng2025DiffusionTW,Yao2025ReconstructionVG}.

In this paper, we propose a different paradigm for diffusion models with pixel-exact reconstruction: rather than modifying diffusion models to facilitate pixel-space generation, \textbf{we construct a lossless transformation that maps pixels into a representation more amenable to diffusion modeling}. Specifically, we introduce the Lossless LDM (LLDM), which reorganizes image information into a diffusion-friendly latent representation while preserving pixel-exact reconstruction. Unlike existing pixel-space diffusion approaches, LLDM requires no modifications to the diffusion model architecture, training objective, or optimization procedure, enabling direct integration into standard latent diffusion frameworks. We construct the lossless transformation through residual decomposition and uneven latent scaling, and propose semantic rank priority (SRP) as a metric to explain its effectiveness. Experiments on ImageNet~\citep{Deng2009ImageNetAL} class-to-image and Diffusion-Bench~\citep{Leng2026DiffusionBenchOH} text-to-image benchmarks show that LLDM substantially outperforms prior pixel-space diffusion methods and even several lossy VAEs in generation quality. Furthermore, SRP effectively predicts downstream generative performance.

Our contributions are summarized as follows:
\begin{itemize}
\item We introduce LLDM, a new paradigm for diffusion modeling with pixel-exact reconstruction. LLDM maps images into diffusion-friendly representations through a lossless transformation while preserving the standard LDM training pipeline.
\item We develop a practical lossless transformation based on residual decomposition and uneven latent scaling, and propose semantic rank priority (SRP) as a metric to explain its effectiveness.
\item Empirically, LLDM outperforms prior pixel-space diffusion models in generation quality while guaranteeing pixel-exact reconstruction. Furthermore, SRP effectively predicts downstream generative performance.
\end{itemize}

\section{Preliminaries}
\subsection{Pixel Diffusion Models}

Let $x \sim p_{\mathrm{data}}(x)$ denote an image sampled from the data distribution. Pixel diffusion models directly model this distribution in pixel space by progressively perturbing $x$ with Gaussian noise over time $t\in[0,T]$. Given a noise schedule defined by $\alpha_t$ and $\sigma_t$, the marginal forward distribution is
\begin{gather}
q(x_t \mid x)
= \mathcal{N}\left(\alpha_t x, \sigma_t^2 I\right).
\end{gather}

The reverse process is described by a reverse-time stochastic differential equation (SDE)~\citep{Anderson1982ReversetimeDE}. Sampling proceeds by simulating this SDE from $t=T$ to $t=0$~\citep{Song2020ScoreBasedGM}, requiring an estimate of the score function $\nabla_{x_t}\log q_t(x_t)$. Accordingly, a score network $s_\theta(x_t,t)$ is trained to approximate this function using the denoising score matching (DSM) objective~\citep{Vincent2011ACB}:
\begin{gather}
\mathcal{L}_{\mathrm{DSM}}
=
\mathbb{E}
\left[
\left\|
s_\theta(x_t,t)
-
\nabla_{x_t}\log q(x_t\mid x)
\right\|_2^2
\right].
\label{eq:dsm}
\end{gather}

Despite its conceptual simplicity, modeling natural images directly in pixel space remains challenging. Recent methods address this through improved training objectives and regularization~\citep{li2026back,Ma2026PixelGenIP,Baade2026LatentFR}, or architectures tailored to pixel-space generation~\citep{Yu2025PixelDiTPD,Teng2023RelayDU,Ma2025DeCoFP}. In contrast, our approach preserves both the diffusion objective and model architecture, instead mapping images into a lossless representation that is easier for diffusion models to learn.

\begin{figure*}[thb]
    \centering
    \includegraphics[width=0.9\linewidth]{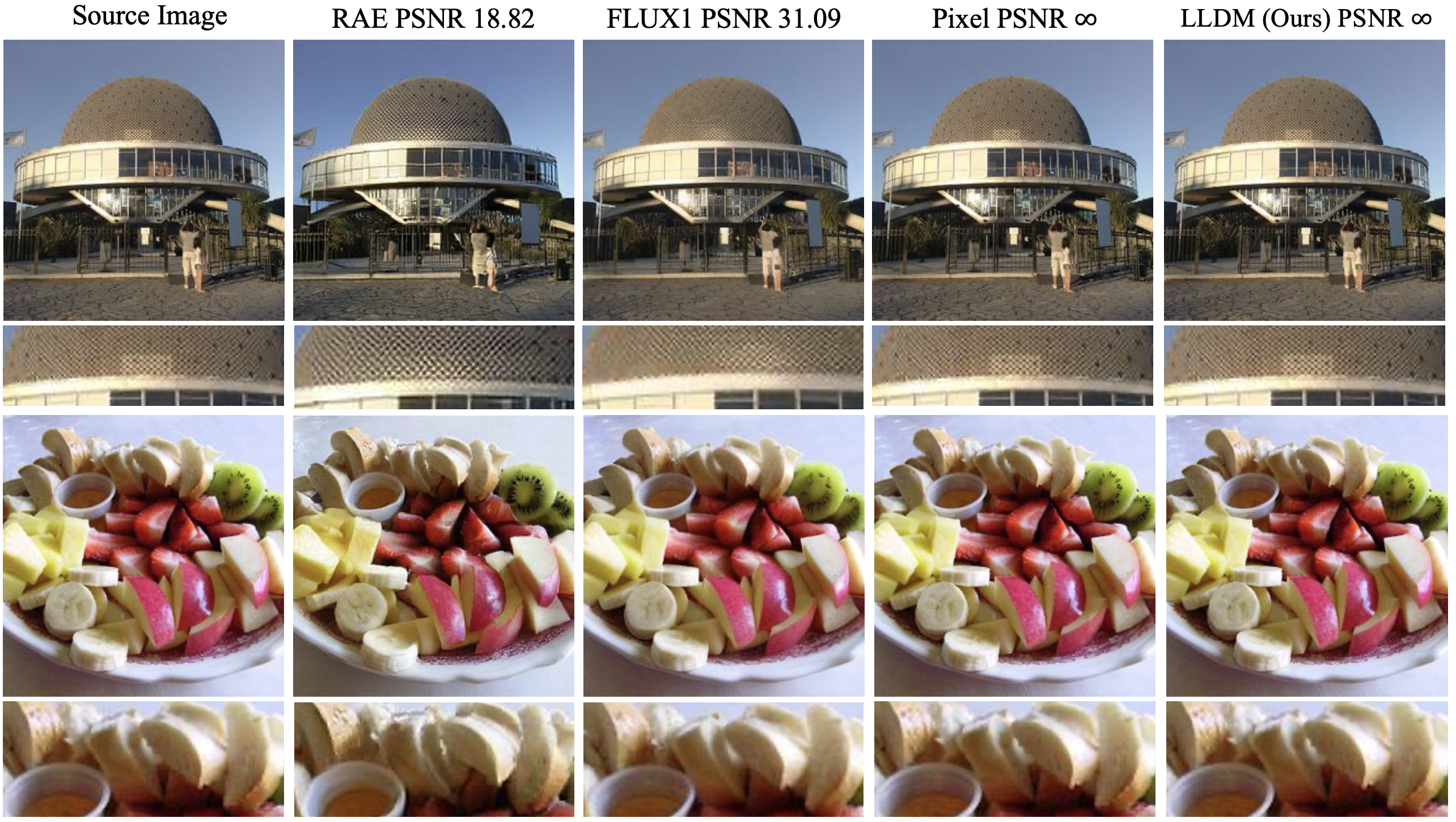}
    \caption{Reconstruction comparison across different VAEs. Both pixel-space diffusion models and our LLDM achieve pixel-exact reconstruction with infinite PSNR.}
\label{fig:rec}
\end{figure*}

\subsection{Latent Diffusion Models}

Latent diffusion models (LDMs)~\citep{rombach2022high} address the challenge of modeling high-dimensional pixel distributions by performing diffusion in a learned latent space. Specifically, an LDM first trains a lossy variational autoencoder (VAE)~\citep{Kingma2013AutoEncodingVB} to map images into compact latent representations. Given an image $x$, an encoder $f(\cdot)$ maps it to a latent $z$, on which the diffusion model is trained. During generation, the sampled latent $z$ is mapped back to pixel space by a decoder $g(\cdot)$:
\begin{gather}
\textrm{Encode:}\quad z = f(x),
\qquad
\textrm{Decode:}\quad \hat{x} = g(z).
\end{gather}

By performing diffusion in a compact latent space, LDMs simplify generative modeling and enable efficient, high-quality image generation. However, conventional LDMs employ lossy VAEs, where $g(f(x)) \neq x$ in general. Information discarded during encoding cannot be recovered, preventing pixel-exact reconstruction. This limitation is particularly undesirable for reconstruction-sensitive applications, such as image editing and restoration, where preserving the original image information is critical.

\section{Lossless Latent Diffusion Model}

In this section, we explore how to construct a lossless representation for latent diffusion. Our goal is to transform data $x$ into a representation that enables pixel-exact reconstruction while remaining as easy for diffusion models to learn as the latent space of a lossy VAE. We start with a simple lossless construction that yields poor generation quality, then progressively introduce designs to improve latent diffusibility.

\subsection{Lossless Transformation via Residual Decomposition}

We begin with a simple lossless transformation. Given a pretrained lossy VAE with encoder $f(\cdot)$ and decoder $g(\cdot)$, we preserve the information lost during reconstruction through a residual:
\begin{gather}
r = x - g(z), \qquad z = f(x).
\end{gather}

\begin{figure*}[thb]
    \centering
    \includegraphics[width=\linewidth]{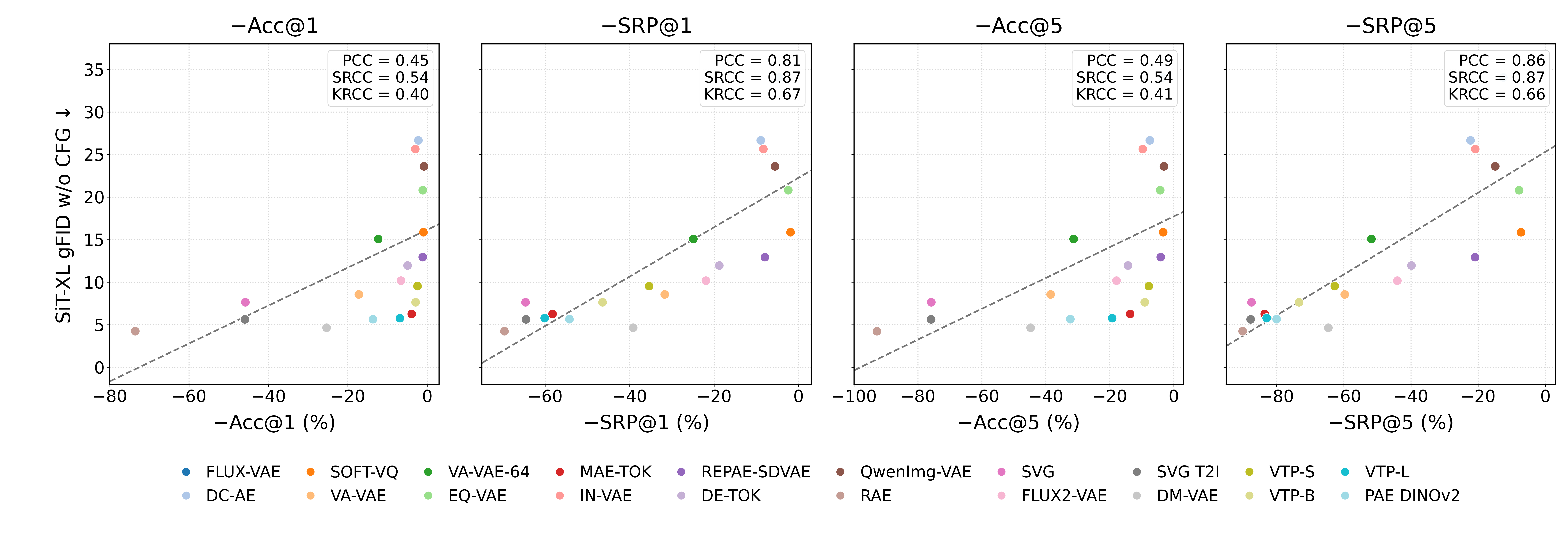}
    \caption{Correlation of ImageNet classification accuracy (Acc@1, Acc@5) and our semantic rank priority (SRP@1, SRP@5) with SiT-XL gFID without CFG. SRP exhibits stronger correlations with gFID across PCC, SRCC, and KRCC. @1, @5 denotes top 1, top 5 accuracy.}
    \label{fig:corr}
\end{figure*}
We construct a new latent representation $\hat{z}=[z,r]$ by concatenating the lossy latent $z$ and residual $r$ along the channel dimension, after spatially rearranging $r$ to match the resolution of $z$. The encoding and decoding processes are:
\begin{gather}
\textrm{Encode:}\quad
z=f(x),\quad r=x-g(z),\quad \hat{z}=[z,r],
\notag\\
\textrm{Decode:}\quad
[z,r]=\hat{z},\quad x=g(z)+r.
\end{gather}

By construction, the original image can be recovered exactly from $\hat{z}$ without reconstruction error. This provides a simple way to convert any pretrained lossy VAE into a lossless representation without modifying its encoder or decoder. However, losslessness does not guarantee suitability for diffusion. As shown in Table~\ref{tab:abl}, modeling $\hat{z}=[z,r]$ is substantially more challenging than modeling the original lossy latent $z$, resulting in significantly worse generation quality. This motivates us to reorganize the residual information into a more diffusion-friendly representation.

\subsection{Lossless Transformation via Uneven Latent Scaling}

A common practice in latent diffusion models is to normalize latent channels to comparable scales before diffusion training, which we refer to as even latent scaling. However, our lossless latent $\hat{z}=[z,r]$ has a heterogeneous structure. For a VAE with high reconstruction quality, the lossy latent $z$ captures most perceptually and semantically important image content, while $r$ preserves the remaining information required for pixel-exact reconstruction. Treating these components equally during diffusion training may therefore be suboptimal.

One solution is to assign different loss weights or training schedules to different latent components~\citep{Fan2025ThePH,Chen2025DCAE1A}. However, this requires modifying the diffusion training objective or procedure. Instead, we preserve the standard LDM pipeline and modify only the latent transformation. Specifically, we propose uneven latent scaling, which scales the residual by a factor $\gamma$ before diffusion training:
\begin{gather}
    \textrm{Encode:}\quad z = f(x),\quad \bar{r} = \gamma(x - g(z)),\quad \hat{z} = [z, \bar{r}], \notag \\
    \textrm{Decode:}\quad [z,\bar{r}] = \hat{z},\quad x = g(z)+\frac{\bar{r}}{\gamma}.
\end{gather}

Choosing $\gamma<1$ reduces the residual scale relative to the lossy latent $z$, encouraging the diffusion model to prioritize the dominant image representation while preserving all information required for exact reconstruction. Importantly, this scaling is invertible, as the original residual can be recovered by dividing by $\gamma$ during decoding. Thus, uneven latent scaling requires no modifications to the diffusion architecture or training objective.

\subsection{Why Not Normalizing Flows?}

Normalizing flows~\citep{Dinh2014NICENI,Dinh2016DensityEU} use invertible transformations to map data into simpler distributions. In principle, an invertible neural network $h_{\theta}(\cdot)$ can also serve as an autoencoder:
\begin{gather}
    \textup{Encode:}\quad \hat{z} = h_{\theta}(x),
    \qquad
    \textup{Decode:}\quad x = h^{-1}_{\theta}(\hat{z}).
\end{gather}
However, as shown in Table~\ref{tab:flowloss}, maximum-likelihood training of normalizing flows yields worse generation quality than pixel-space diffusion. Aligning flow representations with DINO features mitigates this issue but sacrifices invertibility~\citep{Chen2025FlowingBI}. Like normalizing flows, our lossless transformation uses basic operations such as channel splitting, concatenation, and scaling. Nevertheless, learning an invertible transformation optimized end-to-end for diffusion modeling remains an open challenge.

\begin{table}[thb]
\caption{Ablation study of the learned normalizing flow $h_{\theta}(\cdot)$. Introducing a normalizing flow does not improve generation quality.}\label{tab:flowloss}
\centering
\resizebox{0.7\linewidth}{!}{
\begin{tabular}{@{}lcc@{}}
\toprule
 Flow Model & PSNR$\uparrow$ & gFID$\downarrow$ \\ \midrule
- & $\infty$ & 104.29 \\
NICE Flow & $\infty$ & 181.92 \\
RealNVP Flow  & NaN & NaN \\
NICE Flow + VF loss & 56 & 103.77 \\
\bottomrule
\end{tabular}
}
\end{table}

\begin{table}[thb]
\centering
\caption{Relationship between classification accuracy and gFID. Linear classification accuracy fails to capture the effects of uneven latent scaling, whereas our semantic rank priority (SRP) metric correlates strongly with generation quality. @1 denotes top 1 accuracy.}
\label{tab:srp}
\resizebox{\linewidth}{!}{
\begin{tabular}{@{}lccc@{}}
\toprule
Latent Space                     & Acc@1 & SRP@1 (k=100, Ours) & gFID$\downarrow$ \\ \midrule
Pixel $x$ & 0.013 & 0.018 & 104.29           \\
Res. $[z,r]$ & 0.066 & 0.062 & 52.31            \\
Res. + Scale $[z,\bar{r}]$ & 0.064 & 0.167               & 11.66            \\
FLUX2 $z$                        & 0.190 & 0.182               & 10.18            \\ \bottomrule
\end{tabular}
}
\end{table}

\subsection{Understanding LLDM Through Semantic Rank Priority}

To understand why LLDM works and how lossless transformations improve latent diffusibility, we investigate the properties that make latent representations suitable for diffusion. Prior studies have shown that latent diffusibility strongly correlates with semantic separability, measured by linear classification accuracy on ImageNet~\citep{yao2025towards,Zhong2026DiffusingIT}. However, linear classification accuracy cannot explain the effectiveness of LLDM, as it is invariant to invertible linear transformations. Consequently, it remains unchanged under uneven latent scaling.

To address this limitation and better understand the effectiveness of LLDM, we refine the linear classification accuracy metric and propose semantic rank priority (SRP). Specifically, we consider only the top $k$ principal components of the latent representation. We first compute the covariance matrix $\Sigma$ of the latent representation $\hat{z}$ and then obtain the matrix $V_k$ consisting of the eigenvectors corresponding to the $k$ largest eigenvalues of $\Sigma$:
\begin{gather}
    \Sigma = \textrm{Cov}[\hat{z}] = V\Lambda V^T,\notag \\
    \Sigma_k = V_k\Lambda_k V_k^T,\notag \\
    \textrm{where }\Lambda_k = \textrm{Diag}(\lambda_1,\ldots,\lambda_k),\quad V_k=(v_1,\ldots,v_k).
\end{gather}
Instead of evaluating linear classification accuracy on the full latent representation $\hat{z}$, we evaluate it on the latent representation projected onto the rank-$k$ principal subspace:
\begin{gather}
    \hat{z}_k = V_k^T \hat{z}.
\end{gather}

When all principal components are retained, linear classification accuracy on $\hat{z}_k$ is equivalent to that on the original latent representation $\hat{z}$. However, when $k$ is small, the classification accuracy becomes sensitive to linear transformations of the latent representation, such as uneven latent scaling. 

As shown in Table~\ref{tab:srp}, the classification accuracy measured by SRP with $k=100$ aligns well with diffusion sample quality, as measured by gFID. In contrast, conventional linear classification accuracy fails to capture the effect of uneven latent scaling. Furthermore, SRP provides insights into the effectiveness of individual components of LLDM. As shown in Table~\ref{tab:srp}, residual decomposition improves classification accuracy when only the top 100 principal components of the latent representation $\hat{z}$ are retained. This suggests that residual decomposition increases the priority of semantically relevant information in the latent space. Uneven latent scaling further improves classification accuracy by a substantial margin by down-weighting the residual component.

\begin{table*}[htb]
\caption{Ablation study of LLDM components. Both residual decomposition and uneven latent scaling improve generation quality. \textcolor{gray}{Gray} indicates lossy reconstruction (PSNR$\neq\infty$).}
\label{tab:abl}
\centering
\resizebox{0.8\linewidth}{!}{
\begin{tabular}{@{}lcccc@{}}
\toprule
Method & Transform & Residual Processing  & PSNR$\uparrow$ & gFID w/o CFG$\downarrow$ \\ \midrule
\textcolor{gray}{FLUX2} & \textcolor{gray}{$\hat{z} = z$} & - & \textcolor{gray}{34.33} & \textcolor{gray}{10.18} \\
\textcolor{gray}{FLUX2 + Padding} & \textcolor{gray}{$\hat{z} = [z,0]$} & - &\textcolor{gray}{34.33} & \textcolor{gray}{11.71} \\ 
\midrule
Pixel & $\hat{z} = \textup{PixelUnshuffle}(x)$ & - & $\infty$ & 104.29 \\
FLUX2 + Residual & $\hat{z} = [z, r]$ & - & $\infty$ & 52.31 \\
FLUX2 + Residual & $\hat{z} = [z, \bar{r}]$ & - & $\infty$ & 48.37 \\ 
FLUX2 + Residual & $\hat{z} = [z, \bar{r}]$ & Dropout (UAE) & $\infty$ & 23.33 \\
FLUX2 + Residual & $\hat{z} = [z, \bar{r}]$ & Scale (Ours) & $\infty$ & 11.66 \\
\bottomrule
\end{tabular}
}
\end{table*}

\begin{table*}[]
\caption{Ablation study of the scaling parameter $\gamma$. The optimal $\gamma$ varies across base lossy VAEs with different reconstruction quality (PSNR). \textbf{Bold}: Best result.}
\label{tab:ablscale}
\centering
\resizebox{0.7\linewidth}{!}{
\begin{tabular}{@{}cc|cc|cc|cc@{}}
\toprule
Scale $\gamma$ & PSNR & \makecell{Base VAE \\ (Base PSNR)} & gFID & \makecell{Base VAE \\ (Base PSNR)} & gFID & \makecell{Base VAE \\ (Base PSNR)} & gFID \\ \midrule
1.0 & \multirow{4}{*}{$\infty$} & \multirow{4}{*}{\makecell{RAE \\ (18.82)}} & \textbf{38.52} & \multirow{4}{*}{\makecell{VTP-L \\ (26.85)}} & 200.01 & \multirow{4}{*}{\makecell{FLUX2 \\ (34.33)}} & 52.31 \\
0.1 &  &  & 67.70 &  & \textbf{12.91} &  &  19.89\\
0.01 &  &  & 177.38  &  & 15.17 &  & \textbf{11.66} \\
0.001 &  &  & 204.33 &  & 166.71 &  & 11.79 \\ \bottomrule
\end{tabular}
}
\end{table*}

\begin{table}[]
\caption{Ablation study of the base lossy VAE. Despite achieving better generation quality, some lossy VAEs have lower reconstruction fidelity (PSNR), which can degrade the final generation performance of LLDM.}
\label{tab:ablbase}
\centering
\resizebox{0.9\linewidth}{!}{
\begin{tabular}{@{}cllcc@{}}
\toprule
Base VAE & base PSNR & base gFID & PSNR$\uparrow$ & gFID$\downarrow$ \\ \midrule
RAE      & 18.82 & 4.25 & $\infty$ & 38.52 \\
VTP-L    & 26.85 & 5.87 & $\infty$ & 12.91 \\
FLUX2    & 34.33& 10.18 & $\infty$ & 11.66 \\ \bottomrule
\end{tabular}
}
\end{table}

\section{Experimental Results}

\subsection{Experimental Setups}

\textbf{Datasets and metrics.}
For class-to-image experiments, we use ImageNet at $256\times256$ resolution, training diffusion models on the training split and evaluating on the validation split. We report peak signal-to-noise ratio (PSNR) for reconstruction fidelity and generation FID (gFID), Inception Score (IS), precision, and recall for generation quality, following \citet{Dhariwal2021DiffusionMB}. LLDM achieves pixel-exact reconstruction by construction. For text-to-image experiments, we evaluate on Diffusion-Bench at $256\times256$ resolution~\citep{Leng2026DiffusionBenchOH}, using GenEval, DPG-Bench, and GenAI-Bench. To evaluate SRP as a standalone metric, we use the generation results benchmarked by \citet{Xu2026MakingRF}.

\textbf{Baselines and diffusion models.}
We compare LLDM against three groups of baselines. First, we consider recent pixel-space diffusion methods, including pixel-space REPA~\citep{Yu2024RepresentationAF}, JiT~\citep{li2026back}, PixelGen~\citep{Ma2026PixelGenIP}, and Latent Forcing~\citep{Baade2026LatentFR}. These methods enable pixel-space diffusion through modifications to model architectures, representations, or training objectives. Second, we include widely used lossy VAEs without explicit diffusion-oriented representation alignment: SD-VAE, IN-VAE, QwenImage-VAE, FLUX-VAE, and SD3-VAE~\citep{rombach2022high,Yao2025ReconstructionVG,flux2024,esser2024scaling,wu2025qwen}. Third, we compare against diffusion-optimized lossy representations, including VA-VAE, MAE-TOK, VTP-L, and FLUX2-VAE~\citep{Yao2025ReconstructionVG,chen2025softvq,chen2025aligning,yang2025latent}. These baselines enable comparisons with both information-preserving pixel-space diffusion and lossy representations optimized for diffusion.

For each latent representation, we train SiT-XL~\citep{Ma2024SiTEF} to evaluate downstream generation performance. Unless otherwise specified, all diffusion models are trained for 40 epochs using Adam under the same protocol. We report generation quality both with and without classifier-free guidance (CFG)~\citep{Ho2022ClassifierFreeDG}.

\begin{table*}[thb]
\caption{Comparison of PSNR and class-to-image generation performance. LLDM achieves generation quality comparable to lossy VAEs while substantially outperforming other lossless methods (PSNR$=\infty$). \textcolor{gray}{Gray}: Lossy reconstruction (PSNR$\neq\infty$). \textbf{Bold}: Best among lossless methods.}
\label{tab:quant1}
\centering
\resizebox{0.8\linewidth}{!}{
\begin{tabular}{@{}lccccccccc@{}}
\toprule
\multirow{2}{*}{Method} & \multirow{2}{*}{PSNR$\uparrow$} & \multicolumn{4}{c}{SiT/XL w/o CFG}                                                                      & \multicolumn{4}{c}{SiT/XL w/ CFG}                                                                       \\ \cmidrule(lr){3-6}\cmidrule(lr){7-10}
                        &                       & \multicolumn{1}{c}{gFID$\downarrow$} & \multicolumn{1}{c}{IS$\uparrow$} & \multicolumn{1}{c}{Pre.$\uparrow$} & \multicolumn{1}{c}{Rec.$\uparrow$} & \multicolumn{1}{c}{gFID$\downarrow$} & \multicolumn{1}{c}{IS$\uparrow$} & \multicolumn{1}{c}{Pre.$\uparrow$} & \multicolumn{1}{c}{Rec.$\uparrow$} \\ \midrule
SD-VAE & \textcolor{gray}{25.91} & \textcolor{gray}{26.02} & \textcolor{gray}{53.90} & \textcolor{gray}{0.60} & \textcolor{gray}{0.62} & \textcolor{gray}{6.31} & \textcolor{gray}{178.64} & \textcolor{gray}{0.84} & \textcolor{gray}{0.43} \\
SD3-VAE & \textcolor{gray}{29.57} & \textcolor{gray}{26.56} & \textcolor{gray}{50.39} & \textcolor{gray}{0.59} & \textcolor{gray}{0.62} & \textcolor{gray}{7.32} & \textcolor{gray}{189.94} & \textcolor{gray}{0.87} & \textcolor{gray}{0.36} \\
FLUX1 & \textcolor{gray}{31.09} & \textcolor{gray}{33.50} & \textcolor{gray}{41.57} & \textcolor{gray}{0.54} & \textcolor{gray}{0.63} & \textcolor{gray}{8.45} & \textcolor{gray}{160.82} & \textcolor{gray}{0.84} & \textcolor{gray}{0.38} \\
QW-VAE & \textcolor{gray}{30.27} & \textcolor{gray}{23.43} & \textcolor{gray}{55.86} & \textcolor{gray}{0.61} & \textcolor{gray}{0.61} & \textcolor{gray}{7.27} & \textcolor{gray}{180.18} & \textcolor{gray}{0.85} & \textcolor{gray}{0.41} \\
EQ-VAE & \textcolor{gray}{24.92} & \textcolor{gray}{20.79} & \textcolor{gray}{63.14} & \textcolor{gray}{0.65} & \textcolor{gray}{0.60} & \textcolor{gray}{6.34} & \textcolor{gray}{201.39} & \textcolor{gray}{0.88} & \textcolor{gray}{0.39} \\
MAE-TOK & \textcolor{gray}{23.52} & \textcolor{gray}{6.36} & \textcolor{gray}{132.73} & \textcolor{gray}{0.75} & \textcolor{gray}{0.60} & \textcolor{gray}{3.83} & \textcolor{gray}{199.88} & \textcolor{gray}{0.81} & \textcolor{gray}{0.54} \\
VA-VAE & \textcolor{gray}{26.31} & \textcolor{gray}{9.68} & \textcolor{gray}{100.79} & \textcolor{gray}{0.73} & \textcolor{gray}{0.57} & \textcolor{gray}{4.20} & \textcolor{gray}{238.73} & \textcolor{gray}{0.86} & \textcolor{gray}{0.45} \\
RAE & \textcolor{gray}{18.82} & \textcolor{gray}{4.25} & \textcolor{gray}{192.71} & \textcolor{gray}{0.85} & \textcolor{gray}{0.51} & \textcolor{gray}{4.85} & \textcolor{gray}{256.41} & \textcolor{gray}{0.88} & \textcolor{gray}{0.45} \\
VTP-L & \textcolor{gray}{26.85} & \textcolor{gray}{5.87} & \textcolor{gray}{140.61} & \textcolor{gray}{0.78} & \textcolor{gray}{0.59} & \textcolor{gray}{3.48} & \textcolor{gray}{211.17} & \textcolor{gray}{0.83} & \textcolor{gray}{0.53} \\
FLUX2-VAE & \textcolor{gray}{34.33} & \textcolor{gray}{10.18} & \textcolor{gray}{96.68} & \textcolor{gray}{0.72} & \textcolor{gray}{0.61} & \textcolor{gray}{4.16} & \textcolor{gray}{229.93} & \textcolor{gray}{0.87} & \textcolor{gray}{0.49} \\ \midrule
Pixel & $\infty$ & 104.29 & 13.94 & 0.59 & 0.02 & 26.29 & 94.05 & 0.70 & 0.02 \\
Pixel + REPA & $\infty$ &  64.52 & 25.43 & 0.61 & 0.05 & 14.82 & 200.35 & 0.68 & 0.07 \\
JiT & $\infty$ & 114.26 & 15.22 & 0.60 & 0.02 & 24.17 & 103.00 & 0.71 & 0.03 \\
PixelGen & $\infty$ & 63.74 & 25.61 & \textbf{0.65} & 0.06 & 18.95 & \textbf{223.21} & 0.52 & 0.04 \\
LLDM (Ours) & $\infty$ & \textbf{11.64} & \textbf{93.49} & \textbf{0.65} & \textbf{0.54} & \textbf{3.75} & 220.38 & \textbf{0.72} & \textbf{0.55} \\ \bottomrule
\end{tabular}
}
\end{table*}

\begin{table*}[t]
\centering
\begin{minipage}[t]{0.60\textwidth}
\centering
\caption{Comparison of PSNR and text-to-image generation performance. LLDM achieves generation quality comparable to lossy VAEs while substantially outperforming other lossless methods (PSNR$=\infty$). \textcolor{gray}{Gray}: Lossy reconstruction (PSNR$\neq\infty$). \textbf{Bold}: Best among lossless methods.}
\label{tab:t2i}
\vspace{2mm}
\resizebox{\linewidth}{!}{%
\begin{tabular}{@{}lcccc@{}}
\toprule
 & PSNR$\uparrow$ & GenEval$\uparrow$ & DPG-Bench$\uparrow$ & Genai-Bench$\uparrow$ \\
\midrule
RAE & \textcolor{gray}{18.82} & \textcolor{gray}{0.68} & \textcolor{gray}{0.82} & \textcolor{gray}{0.70} \\
SD-VAE & \textcolor{gray}{26.02} & \textcolor{gray}{0.58} & \textcolor{gray}{0.80} & \textcolor{gray}{0.69} \\
SD3-VAE & \textcolor{gray}{29.57} & \textcolor{gray}{0.64} & \textcolor{gray}{0.81} & \textcolor{gray}{0.70} \\
QW-VAE & \textcolor{gray}{30.27} & \textcolor{gray}{0.61} & \textcolor{gray}{0.80} & \textcolor{gray}{0.70} \\
FLUX1 & \textcolor{gray}{31.09} & \textcolor{gray}{0.56} & \textcolor{gray}{0.78} & \textcolor{gray}{0.68} \\
FLUX2 & \textcolor{gray}{34.33} & \textcolor{gray}{0.67} & \textcolor{gray}{0.83} & \textcolor{gray}{0.71} \\
\midrule
Pixel & $\infty$ & 0.50 & 0.75 & 0.64 \\
JiT & $\infty$ & 0.52 & 0.78 & 0.67 \\
PixNerd & $\infty$ & 0.48 & 0.78 & 0.64 \\
PixelGen & $\infty$ & 0.55 & 0.80 & 0.68 \\
LLDM (Ours) & $\infty$ & \textbf{0.67} & \textbf{0.82} & \textbf{0.72} \\
\bottomrule
\end{tabular}%
}
\end{minipage}%
\hfill%
\begin{minipage}[t]{0.375\textwidth}
\centering
\caption{Correlation of VAE metrics to diffusion generation results. SRP significantly improves correlation over vanilla linear accuracy using low-rank projection. \textbf{Bold}: Best. @1, @5: top 1, top 5.}
\label{tab:rcorr}
\vspace{2mm}
\resizebox{\linewidth}{!}{%
\begin{tabular}{@{}llll@{}}
\toprule
\multirow{2}{*}{} & \multicolumn{3}{c}{SiT/XL gFID w/o CFG} \\
\cmidrule(l){2-4}
 & \multicolumn{1}{c}{PCC$\uparrow$}
 & \multicolumn{1}{c}{SRCC$\uparrow$}
 & \multicolumn{1}{c}{KRCC$\uparrow$} \\
\midrule
-PSNR & -0.54& -0.54& -0.39\\
LPIPS & -0.54& -0.56& -0.42\\
LDS & 0.14 & 0.43 & 0.30 \\
CDS & 0.01 & 0.01 & 0.00 \\
SEC & -0.13 & -0.01 & 0.34 \\
GMM & 0.29 & 0.43 & 0.30 \\
VIV & 0.32& 0.41& 0.32 \\
-Acc@1 & 0.45 & 0.54 & 0.40 \\
-Acc@5 & 0.49 & 0.54 & 0.41 \\
-SRP@1 (Ours) & 0.81 & \textbf{0.87} & \textbf{0.67} \\
-SRP@5 (Ours) & \textbf{0.86} & \textbf{0.87} & 0.66 \\
\bottomrule
\end{tabular}%
}
\end{minipage}
\end{table*}

\begin{figure*}[thb]
    \centering
    \includegraphics[width=0.8\linewidth]{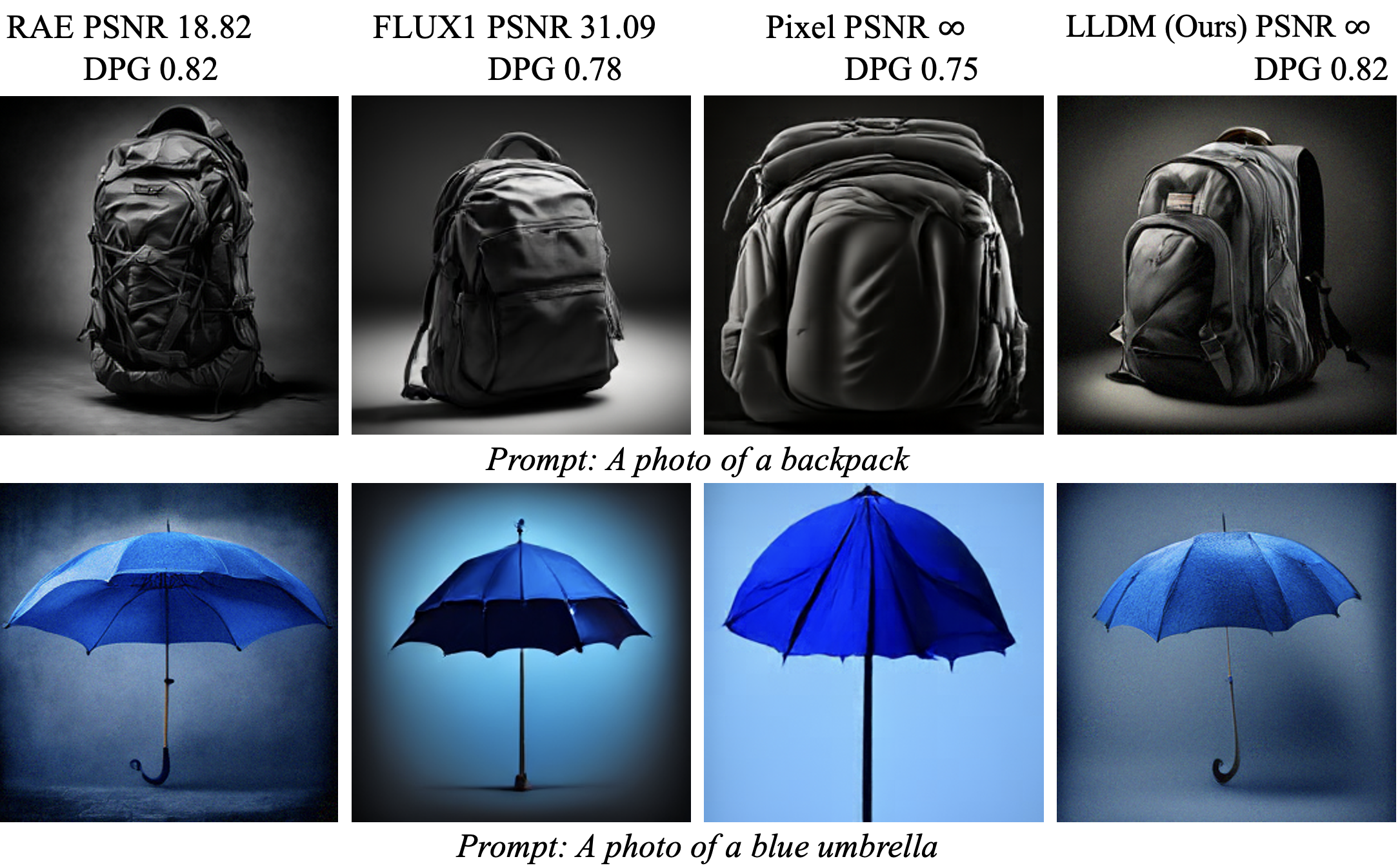}
    \includegraphics[width=0.8\linewidth]{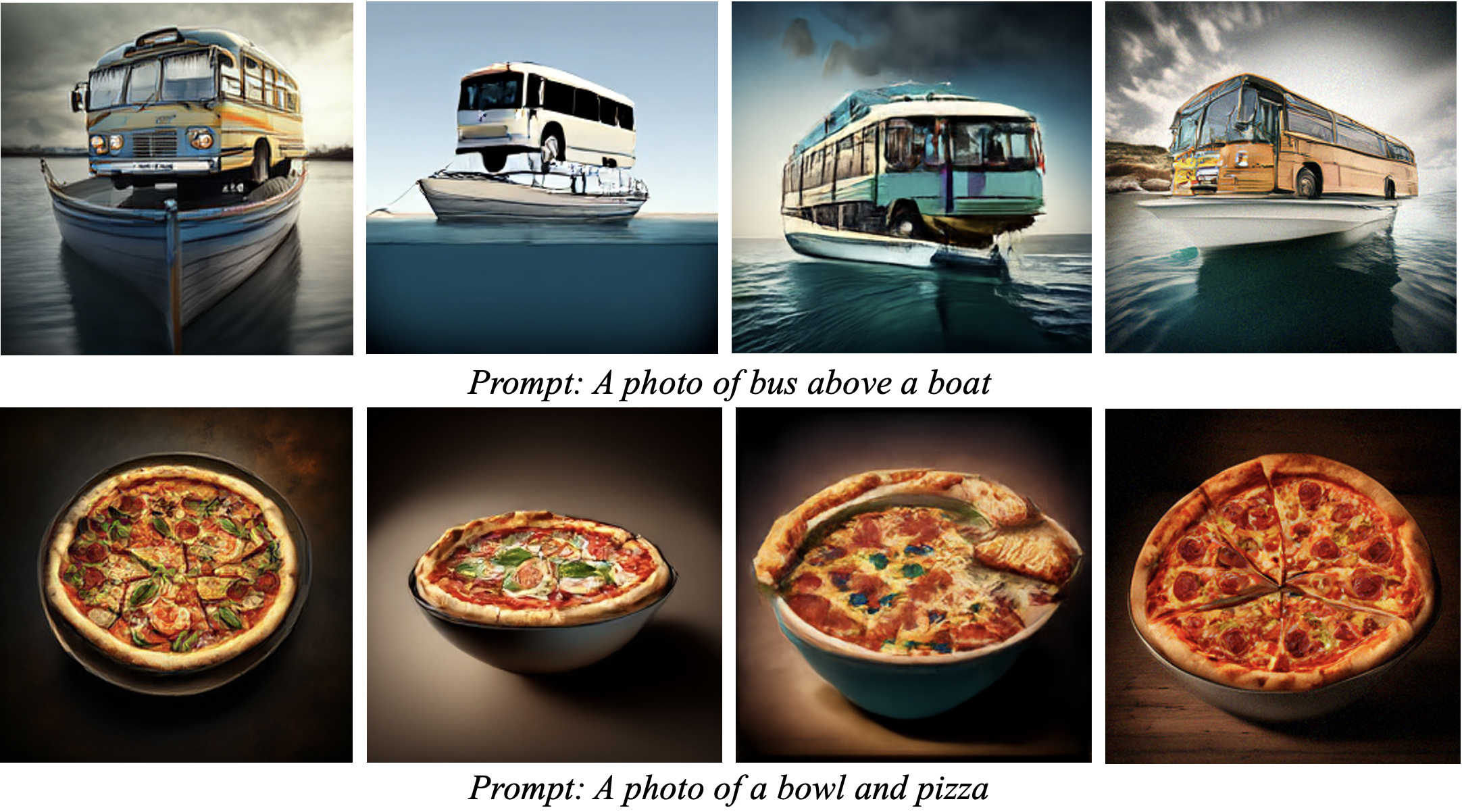}
    \caption{Comparison of generated samples across VAEs. LLDM achieves generation quality comparable to lossy LDMs while substantially outperforming pixel-space diffusion models.}
    \label{fig:gen}
\end{figure*}

\subsection{Main Results}

\textbf{LLDM outperforms pixel-space diffusion models in generation quality.}
As shown in Table~\ref{tab:abl}, LLDM substantially outperforms competing pixel-space diffusion models while achieving pixel-exact reconstruction ($\mathrm{PSNR}=\infty$). These improvements require no modifications to the diffusion architecture or training procedure. LLDM follows the same training pipeline as conventional lossy LDMs, replacing the lossy VAE with our lossless transformation. These results demonstrate that high-quality generation with exact reconstruction does not require specialized pixel-space architectures or objectives. Instead, transforming pixels into a diffusion-friendly, lossless representation can substantially simplify generative modeling.

\textbf{LLDM outperforms many latent diffusion models in both reconstruction and generation.}
Remarkably, LLDM also outperforms many LDMs based on lossy VAEs while preserving all information required for pixel-exact reconstruction. Specifically, LLDM achieves better generation quality than widely used reconstruction-oriented VAEs, including SD-VAE, SD3-VAE, FLUX-VAE, and QwenImage-VAE. Unlike these VAEs, which discard information during encoding, LLDM preserves the input exactly. Moreover, LLDM achieves generation quality comparable to FLUX2-VAE, a competitive diffusion-oriented VAE. These results suggest that information loss is not inherently necessary for strong diffusion performance; rather, how information is organized in the latent space critically determines its diffusibility.

\textbf{SRP effectively predicts diffusion generation quality.}
We further evaluate SRP on a benchmark of 22 open-source lossy VAEs. As shown in Table~\ref{tab:rcorr}, SRP using Acc@5 with rank $k=256$ substantially outperforms conventional classification accuracy (Acc@5). Compared with other VAE metrics, including LDS, CDS, VIV, and GMM loss, SRP also achieves significantly stronger correlations with diffusion gFID, measured by Pearson (PCC), Spearman (SRCC), and Kendall (KRCC) correlation coefficients.

\subsection{Ablation Studies}

\textbf{Effectiveness of lossless transformations.}
In Table~\ref{tab:abl}, we ablate the key components of LLDM to examine how reorganizing image information affects latent diffusibility. Starting from vanilla pixel diffusion (gFID $104.29$), decomposing the image into FLUX2 latents and reconstruction residuals substantially reduces gFID to $52.31$. Applying an invertible normalizing flow to the residual further reduces gFID to $48.37$. Most notably, uneven latent scaling dramatically improves gFID to $11.66$, approaching the original lossy FLUX2-VAE ($10.18$). These results suggest that diffusion difficulty depends not only on the amount of information modeled but also on its organization and scaling. With an appropriate lossless transformation, LLDM preserves all pixel information while approaching the generation quality of a strong lossy representation.

\textbf{Choice of base VAE and latent scaling factor.}
In Table~\ref{tab:ablscale}, we examine the sensitivity of LLDM to the base VAE and uneven scaling factor $\gamma$. Interestingly, the optimal $\gamma$ varies substantially across VAEs and is closely related to reconstruction fidelity. For RAE, which has relatively low reconstruction PSNR, any $\gamma<1.0$ degrades generation quality. For VTP-L, with moderate reconstruction fidelity, $\gamma=0.1$ achieves the best gFID. In contrast, FLUX2-VAE, with high reconstruction fidelity, favors a much smaller $\gamma=0.01$. This trend supports the motivation for uneven latent scaling: as reconstruction fidelity improves, less information remains in the residual, allowing diffusion to place less emphasis on it. Conversely, for highly lossy VAEs, aggressively downscaling the residual suppresses important image information and degrades generation quality.

\section{Related Works}

\textbf{Latent Diffusion Models.}
Latent diffusion models (LDMs)~\citep{rombach2022high} employ lossy VAEs to map images into compact latent representations, accelerating diffusion convergence compared with pixel-space training. Early LDMs primarily optimized VAEs for reconstruction fidelity~\citep{rombach2022high,esser2024scaling}. Subsequent studies showed that better reconstruction does not necessarily improve downstream generation, motivating diffusion-oriented VAE optimization through joint training~\citep{wehenkel2021diffusion,vahdat2021score,Heek2026UnifiedL}, signal-processing constraints~\citep{Kouzelis2025EQVAEER,Skorokhodov2025ImprovingTD,liu2025delving,fan2025prism}, and foundation-model alignment~\citep{chen2025softvq,Chen2025MaskedAA,zheng2025diffusion,Yao2025ReconstructionVG,ye2025distribution,yang2025latent,yao2025towards,Shi2025LatentDM,Gao2025OneLI}. Despite these advances, lossless transformations that preserve semantic information in diffusion-friendly representations remain underexplored.

\textbf{Pixel Diffusion Models.}
Early image diffusion models were primarily trained directly in pixel space~\citep{Ho2020DenoisingDP,vahdat2021score,Dhariwal2021DiffusionMB}. However, vanilla pixel diffusion converges slowly and struggles with high-resolution generation. Subsequent approaches address these limitations by modifying model architectures or training objectives. For example, \citet{Yu2025PixelDiTPD,Ma2025DeCoFP} improve the DiT architecture~\citep{Peebles2022ScalableTransformers} for pixel-space diffusion, while \citet{Baade2026LatentFR,Ma2026PixelGenIP,li2026back} introduce improved training objectives. Despite these advances, lossless transformations that improve the diffusibility of pixel-space data remain underexplored.

\section{Discussion \& Conclusion}

In this work, we introduce a new paradigm for diffusion modeling with pixel-exact reconstruction. Rather than modifying diffusion architectures or training procedures to handle pixel-space complexity, we propose Lossless Latent Diffusion Models (LLDMs), which transform pixels into lossless, diffusion-friendly representations through residual decomposition and uneven latent scaling. We further introduce semantic rank priority (SRP) to explain the effectiveness of LLDM. Experiments demonstrate that LLDM substantially outperforms existing pixel-space diffusion methods in generation quality without specialized architectures or training objectives. Moreover, SRP exhibits strong correlations with downstream diffusion generation quality.

\section*{LLM Usage Statement}
In this work, we used generative AI tools for proof reading and polishing. We have not used generative AI tools for generating research ideas or generating code. We take responsibility for the final content of this work, including text, claims or artifacts produced with the aid of generative AI.

{
    \small
    \bibliographystyle{ieeenat_fullname}
    \bibliography{main}
}


\end{document}